\documentclass[11pt]{article}
\pdfoutput=1

\usepackage{acl}

\usepackage{times}
\usepackage{latexsym}
\usepackage{algorithm}
\usepackage{algpseudocode}
\usepackage{amsmath}
\usepackage[T1]{fontenc}

\usepackage[utf8]{inputenc}

\usepackage{microtype}

\usepackage{inconsolata}

\usepackage{graphicx}

\usepackage{multirow}

\usepackage{enumitem}
\usepackage{subfig}
\usepackage{float}
\usepackage{caption}
\usepackage[capitalize]{cleveref}
\usepackage{hyperref}
\usepackage{makecell}

\title{CliMedBench: A Large-Scale Chinese Benchmark for Evaluating Medical Large Language Models in Clinical Scenarios}

\author{Zetian Ouyang$^{1}$ Yishuai Qiu$^{1}$, Linlin Wang$^{1}$\thanks{Corresponding author.}, Gerard de Melo$^{2}$, \\ \textbf{Ya Zhang$^{3}$, Yanfeng Wang$^{3}$, Liang He$^{1}$}\\
        \textsuperscript{1}East China Normal University \quad
        \textsuperscript{2}Hasso Plattner Institute\\ \textsuperscript{3}Shanghai Jiao Tong University\\
        \{51265901102,51265901063\}@stu.ecnu.edu.cn, \{llwang,lhe\}@cs.ecnu.edu.cn,\\gdm@demelo.org,  \{ya\_zhang,wangyanfeng\}@sjtu.edu.cn}

\begin{document}
\maketitle
\begin{abstract}
With the proliferation of Large Language Models (LLMs) in diverse domains, there is a particular need for unified evaluation standards in Chinese clinical medical scenarios, where models need to be examined very thoroughly.
We present CliMedBench, a comprehensive benchmark with 14 expert-guided core clinical scenarios specifically designed to assess the medical ability of LLMs across 7 pivot dimensions\footnote{\url{https://github.com/Optifine-TAT/CliMedBench}}. It comprises 33,735 questions derived from real-world medical reports of top-tier tertiary hospitals and authentic examination exercises. The reliability of this benchmark has been confirmed in several ways. Subsequent experiments with existing LLMs have led to the following findings: (i) Chinese medical LLMs underperform on this benchmark, especially where medical reasoning and factual consistency are vital, underscoring the need for advances in clinical knowledge and diagnostic accuracy. (ii) Several general-domain LLMs demonstrate substantial potential in medical clinics, while the limited input capacity of many medical LLMs hinders their practical use. These findings reveal both the strengths and limitations of LLMs in clinical scenarios and offer critical insights for medical research.
\end{abstract}

\section{Introduction}
With the advent of Chinese medical large language models (LLMs) such as  Hua\-tuo\-GPT~\citep{zhang-etal-2023-huatuogpt}, ChatMed~\citep{zhu2023ChatMed}, and BenTsao~\citep{wang2023huatuo}, the potential for these tools in healthcare has expanded considerably~\cite{singhal2023large}. These models are engineered to address intricate medical problems by providing diagnostic assistance and treatment suggestions. Nonetheless, the absence of a comprehensive and systematic evaluation of their performance, which encompasses aspects such as response accuracy, hallucination incidence, and content safety, hampers their integration into clinical practice. Consequently, there is an urgent need for a standardized evaluation benchmark to scrutinize the capabilities and limitations of such medical LLMs.

Developing a practically relevant benchmark is non-trivial. There is a substantial disconnect between current benchmarks for the Chinese language and the realities of medical practice, as such benchmarks are mostly derived from open educational resources~\citep{mbakwe2023chatgpt}. A benchmark based on real-world medical cases offers superior authenticity and heterogeneity, while more accurately mirroring the intricacies encountered in clinical practice. These cases present greater challenge and complexity, leading to a more rigorous assessment of model performance and robustness in practical applications, including clinical decision support, diagnosis, and treatment recommendations. Moreover, benchmarks developed from open resources are susceptible to data contamination issues.

Prevalent medical benchmarks like MedQA~\citep{jin2021disease} and MedMCQA~\citep{pal2022medmcqa} incorporate data from accessible sources such as textbooks, scholarly articles, and qualification examinations. The effectiveness of such evaluation benchmarks is controversial: medical exams are inefficient clinical performance indicators. Large-scale EHR-based benchmarks such as emr-QA~\citep{pampari-etal-2018-emrqa} have addressed the deficiency in clinical QA; however, language discrepancies preclude their direct applicability for evaluating Chinese medical LLMs. Chinese benchmarks, including CMExam~\citep{liu2024benchmarking}, CMB~\citep{wang-etal-2024-cmb}, and MLEC-QA~\citep{li2021mlec}, primarily source their data from exams such as CNMLE and NMLEC. MedBench~\citep{cai2024medbench} uses exam questions and artificially generated EHRs to evaluate the LLMs' exam-solving capabilities in different subdisciplines. Despite their comprehensive analysis, these benchmarks are disconnected from actual medical practice due to their lack of real-world medical case data, and the use of exam-based data raises concerns about data pollution. The need for good benchmarks makes evaluating performance a significant challenge.

To address the limitations of prior research, we introduce the CliMedBench, a robust benchmark comprising 33,735 questions across 14 core medical scenarios assessing LLMs' ability across six dimensions, primarily sourced from authentic cases to align with medical standards and practices. The CliMedBench integrates expertise from Chinese medical practitioners, offering a valid measure for gauging medical linguistic proficiency and cognitive skills in LLMs. We evaluate various general and medical-specific LLMs using this benchmark and perform a comprehensive analysis that sheds light on relevant research avenues to enhance the medical capabilities of LLMs. The main findings on this benchmark are as follows:
\begin{itemize}[label=$\bullet$,leftmargin=*]
    \item Chinese medical LLMs underperform on this benchmark, especially where medical reasoning and factual consistency are vital, underscoring the need for advances in clinical knowledge and diagnostic accuracy.
    \item Several general-domain LLMs demonstrate substantial potential in medical clinics, while the limited input capacity of many medical LLMs hinders their practical use.
    \item The indeterminacy inherent in medical contexts can significantly compromise the accuracy of model-generated responses.
\end{itemize}

\section{The Proposed Benchmark}
\label{benchmark}
\subsection{The Taxonomy of CliMedBench}
\begin{figure*}[htbp]
    \centering
    \includegraphics[width=16cm]{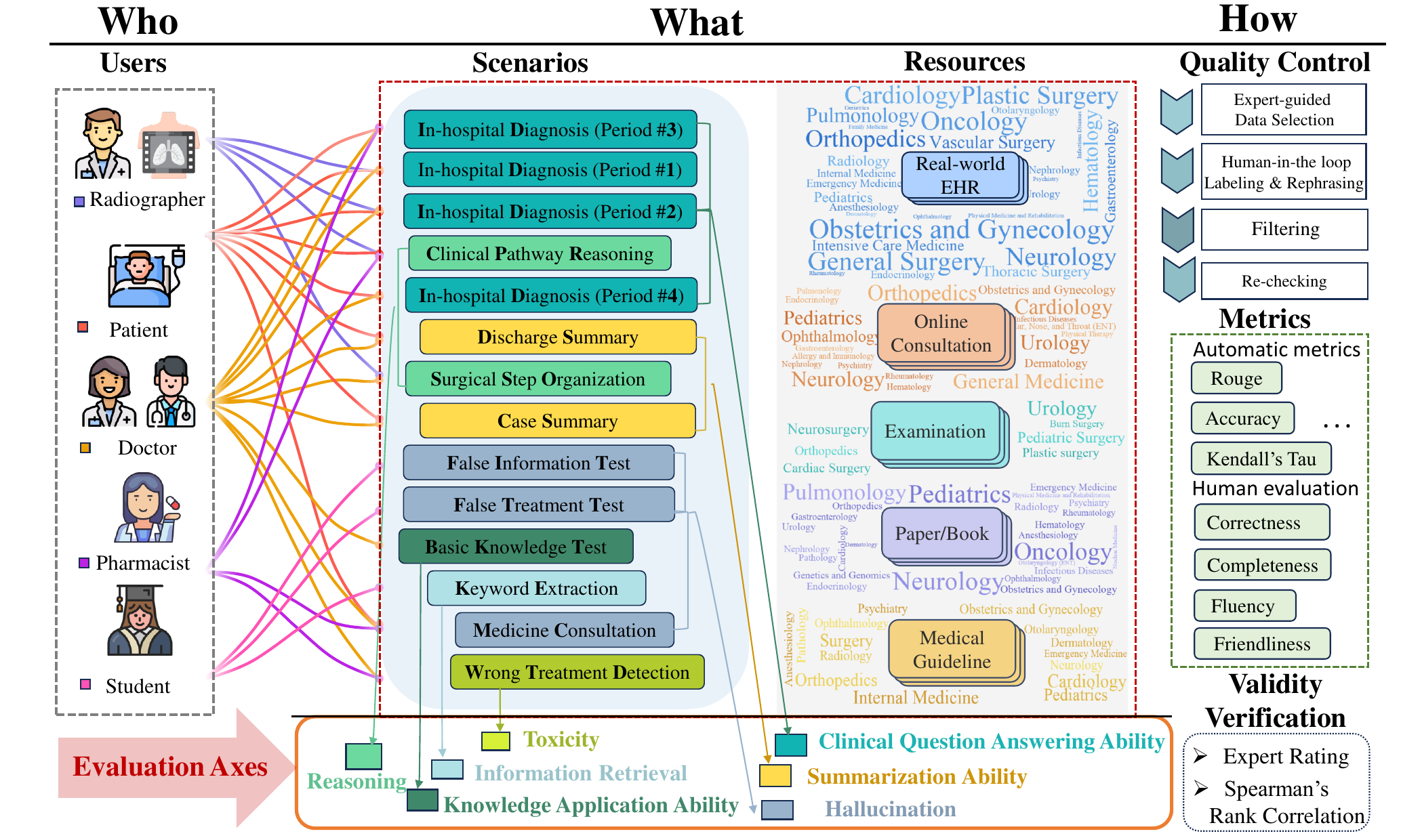}
    \caption{Overview of CliMedBench with ``Who-What-How” taxonomy linking users with core clinical scenarios.} 
    \label{fig: scenarios centered framework}
    \vspace{-8pt}
\end{figure*}
A well-structured taxonomy enables us to conduct a more fine-granular assessment of medical LLMs, while also ascertaining that the evaluation is comprehensive and practically relevant. 
Our taxonomy, designed to maintain the benchmark's applicability and comprehensiveness in real-world clinical scenarios, is based on a categorization that mirrors medical practice and fully covers it, as inspired by~\citet{Liang2022HolisticEO}. As depicted in Figure~\ref{fig: scenarios centered framework}, we build on a ``Who--What--How” scheme to categorize real-world clinical medical practice, providing 14 core clinical scenarios for assessment. Along the ``Who” axis, we distinguish five principal roles in the medical field: the \emph{Radiographer}, \emph{Pharmacist}, \emph{Patient}, \emph{Medical Student}, and \emph{Specialist Doctor}, where doctors encompass attending physicians, surgeons, and other medical specialists. ``What” addresses a broad spectrum of key medical scenarios, covering basic knowledge tests, in-hospital diagnosis, clinical pathway reasoning, case summaries, wrong treatment detection, etc. This allows CliMedBench to evaluate the medical ability of LLMs from seven perspectives, including clinical question answering, knowledge application, reasoning, information retrieval, summarization abilities, hallucination, and toxicity.  

To illustrate the divisions within CliMedBench, we provide an example of the scenario mappings for \textbf{I}n-hospital \textbf{D}iagnoses (ID) in Table~\ref{tab:my_label}. ID is one of the core scenarios in CliMedBench that spans four periods, encapsulating the patient care continuum from admission to discharge, and consider the following scenario descriptions: 
\begin{itemize}
    \item ID \#1 refers to the selection of examinations by healthcare professionals and radiographers.
     \item ID \#2 involves the diagnosis by physicians, integrating examination results with the patient's medical history and additional health data.
     \item In the ID \#3 period, treatment strategies, ranging from pharmacological interventions to surgical procedures, are developed in collaboration with pharmacists and medical staff.
     \item ID \#4 pertains to physicians providing discharge instructions to patients.
\end{itemize}


\begin{table}[htbp]
    \centering
    \begin{tabular}{lc}
\hline
\textbf{Who} & \textbf{What (Task)}\\
\hline
Doctor, Patient &ID\#1\\
Doctor, Radiographer, Patient& ID\#2\\ 
Doctor, Pharmacist, Patient& ID\#2\\
Doctor, Patient &ID\#4\\
\hline
\multicolumn{2}{c}{\textbf{Evaluation axes:} clinical QA ability} \\
\hline
\end{tabular}
    \caption{Example mappings in clinic scenarios. In subsequent sections, tasks will be designated using acronyms formed by the initial letters.}
    \label{tab:my_label}
\end{table}

\subsection{Construction and Statistics}
CliMedBench is derived from real-world Electronic Health Records (EHRs) of top-tier tertiary hospitals in China, supplemented with examination exercises, medical guidelines, textbooks, scholarly articles, and human-annotated online consultations. This corpus spans a multitude of medical specialties, meticulously curated to enhance the diversity of CliMedBench. The EHR data were supplied by our collaborating hospital. The dataset comprises all EHRs that include radiological records from the period January 1, 2023, to March 31, 2023. Examination questions are derived from the NMLEC 2023 Annual Examination, covering 16 subjects within the domains of Surgery and Internal Medicine.

\begin{figure}[h]
    \vspace{-4pt}
    \centering
    \includegraphics[width=6.5cm]{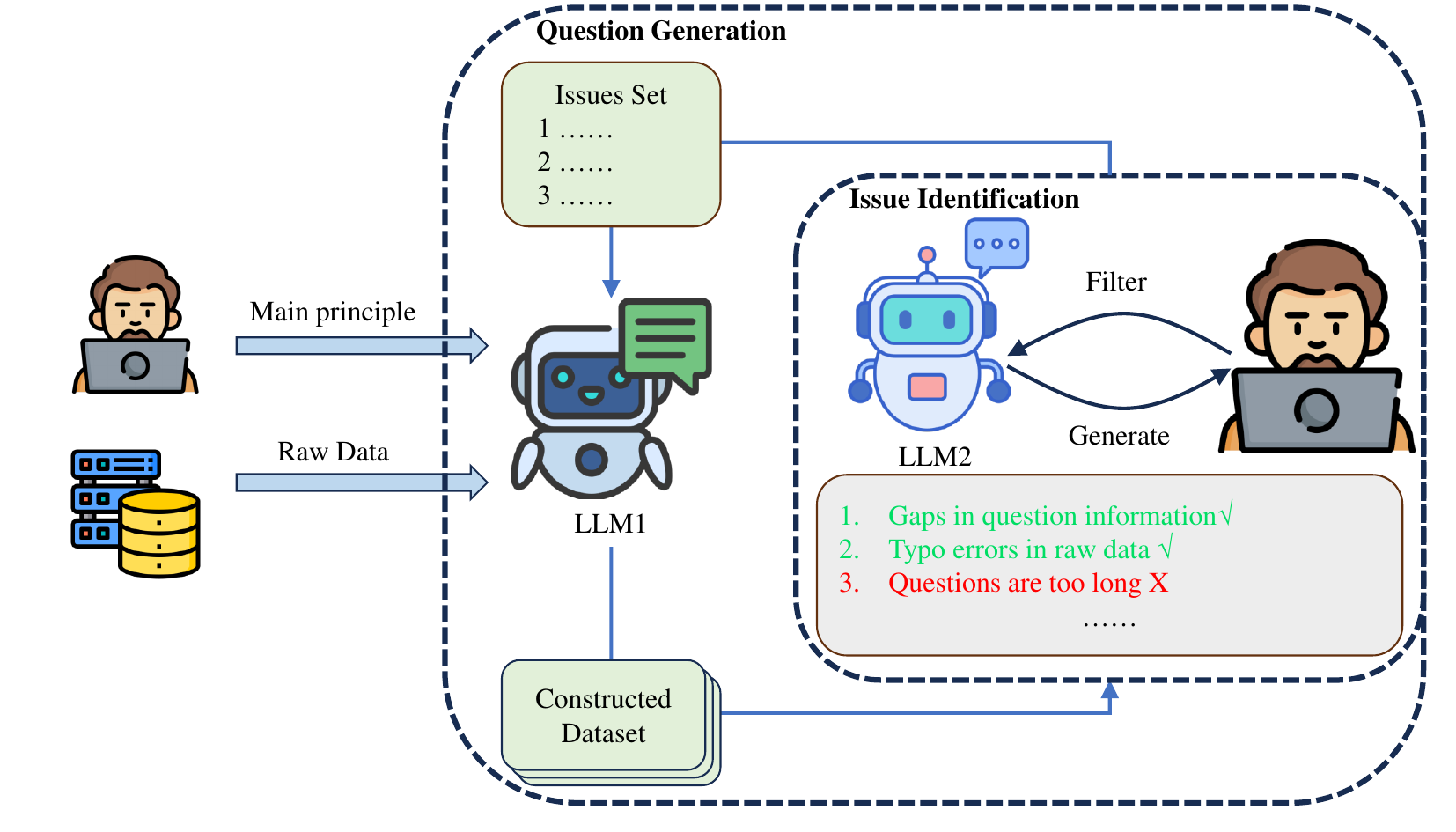}
    \caption{Workflow of collaboration between humans and LLMs for dataset construction.}
    \label{fig: construct dataset with LLMs}
    \vspace{-8pt}
\end{figure}

\begin{figure}[htbp]
    \vspace{-8pt}
    \centering
    \includegraphics[width=8cm]{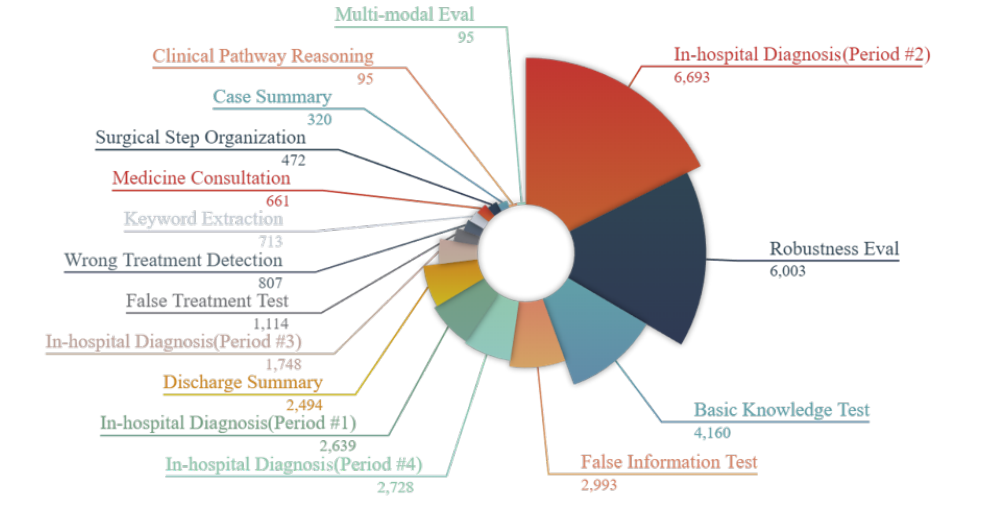}
    \caption{Data distribution of clinical scenarios.}
    \label{fig: dataset capacity}
    \vspace{-4pt}
\end{figure}

Before the dataset construction, all EHRs have been doubly de-identified by ethics committees and experts to make sure no PHI of patients or healthcare professionals is leaked. Then we preprocess the raw data to filter low-quality data and ensure proper formatting by automatic techniques such as regular expression matching and medical entity recognition. Figure~\ref{fig: construct dataset with LLMs} illustrates the workflow of our construction. Medical experts initially develop guidelines based on the content types indicated by doctors' notes in EHRs; these guidelines are used to segment EHRs into sentences, which are then categorized by LLM1 into components for question generation, with QA-pairs formulated by coupling correct answers with distractors sourced from disparate but thematically similar EHR segments. A secondary LLM (LLM2) audits the dataset for inconsistencies or ambiguities, with flagged items evaluated by medical experts to ensure only relevant issues are addressed. Feedback from this evaluation refines LLM1’s processing to mitigate recurrent dataset flaws. The dataset undergoes iterative validation cycles, ensuring a minimum of 90\% of the questions meet quality standards as confirmed by dual expert review. We finally obtained 33,735 instances for 14 core clinical scenarios that are strictly based on doctor's notes and clinical treatment recordings. Figure~\ref{fig: dataset capacity} depicts the data distribution of CliMedBench, encompassing 19 branches of medicine, e.g., neurosurgery and gastroenterology. CliMedBench has three question types, including: 
\begin{itemize}
\setlength{\itemsep}{0pt}
\setlength{\parsep}{0pt}
\setlength{\parskip}{0pt}
    \item Multiple-choice clinical question answering. 
    \item Sequencing questions, e.g., surgical step reordering.  
    \item Open-ended generation, e.g., discharge summary, subjective clinical question answering. 
\end{itemize}

\noindent To confirm the effectiveness of benchmark construction, we employ diverse methodologies to validate CliMedBench, as described in Section~\ref{sec:validate}.

\subsection{Characteristics}
CliMedBench improves over existing benchmarks in several respects: (1) \textbf{Authenticity and Uniqueness}, as it genuinely reflects doctors' practical experience by exclusively using expert-annotated EHRs from top hospitals with up-to-date, authentic insights, while reducing the potential of data contamination. (2) \textbf{Comprehensiveness and Multi-dimensionality}, as it is meticulously designed to align with Chinese clinical practices, encompassing diverse medical disciplines with multimodal information, offering a broad spectrum of evaluation perspectives. (3) \textbf{Practicality}, as it offers 
a novel agent-based Computerized Adaptive Testing approach to guarantee rapid assessment with CliMedBench.


\section{Models and Evaluation Metrics}
To assess the state-of-the-art, we conduct evaluations using CliMedBench of 11 representative LLMs from both the general and medical domains, including OpenAI's GPT series\footnote{https://chat.openai.com} (gpt-3.5-turbo-1106 and gpt-4-1106-preview), ChatGLM3~\cite{zeng2022glm}, ERNIE-Bot 4.0\footnote{https://yiyan.baidu.com}, Xinghuo v3\footnote{https://xinghuo.xfyun.cn}, Qwen max\footnote{https://tongyi.aliyun.com}, Baichuan\footnote{https://github.com/baichuan-inc/Baichuan-13B}, HuaTuoGPT~\citep{zhang-etal-2023-huatuogpt}, BenTsao~\citep{wang2023huatuo}, MedicalGPT~\footnote{https://github.com/shibing624/MedicalGPT}, and ChatMed~\citep{zhu2023ChatMed}. The base model of HuatuoGPT, HuatuoGPT2, Bentsao, and MedicalGPT are Ziya-LLaMA-13B-Pretrain-v1, Baichuan2-13B-Chat, Alpaca-Chinese-7B, and Baichuan-13B-Chat, respectively. We adopt the default or publisher-recommended parameter settings in their published website.
Given the presence of multiple-choice, sequencing, and open-ended generation questions in CliMedBench, we utilize a comprehensive set of metrics. Specifically, we use Accuracy for multiple-choice question answering and Kendall's $\tau$~\cite{kendall1938new} for sequencing questions. For open-ended generation, we combine expert-level human evaluation with 
supplementary automatic evaluation metrics, e.g., ROUGE-1~\cite{lin2004rouge} for discharge summary and SimCSE-based similarity for wrong treatment assessment. For the latter, we first apply fine-tuning to SimCSE~\citep{gao-etal-2021-simcse} using distinct medical documents, then utilize the resulting model to derive sentence vectors, and finally compute the semantic similarity with the reference. 

\section{Main Results}
\begin{figure}[htbp]
  \vspace{-6pt}
  \centering
\includegraphics[width=0.9\linewidth]{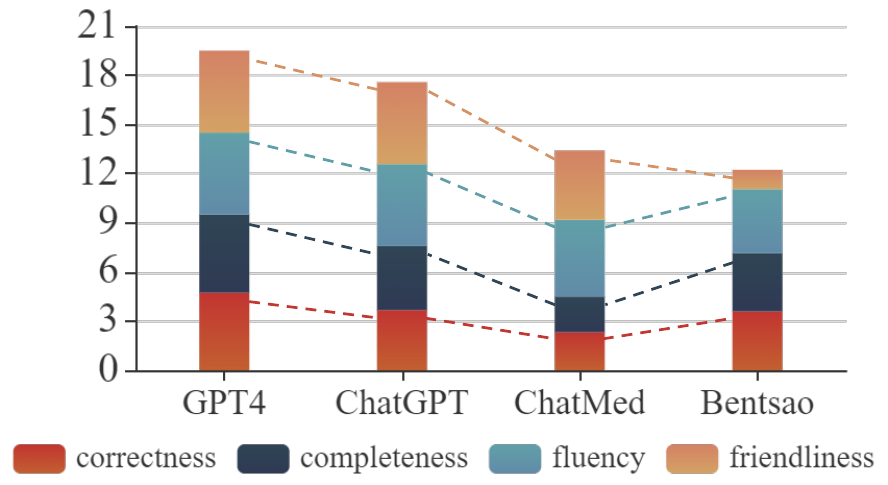}
  \caption{Human evaluation results of four aspects.}
  \label{fig: output human evaluation}
  \vspace{-8pt}
\end{figure}

We conduct an in-depth evaluation of 11 LLMs using CliMedBench, stringently examining their performance across seven pivot dimensions. Corresponding comparisons utilizing automatic metrics are provided in Table~\ref{tab: table1}.
We also engage human experts to assess open-ended generation (WTD and multi-modal report analysis) across four dimensions in Figure~\ref{fig: output human evaluation}, including medical correctness, completeness, fluency, and friendliness. 
 \begin{table*}[h] 
 \centering
 \renewcommand\arraystretch{0.8}
 \begin{tabular}{ll|ccccccc}
 \hline
 \multicolumn{2}{l|}{\textbf{\quad Model}} & \multicolumn{1}{c}{ID\#3} & \multicolumn{1}{c}{ID\#1}& \multicolumn{1}{c}{ID\#2} & \multicolumn{1}{c}{CPR}& \multicolumn{1}{c}{ID\#4}& \multicolumn{1}{c}{DS}& \multicolumn{1}{c}{SSO}\\
 \multicolumn{2}{c|}{} & \multicolumn{1}{c}{\emph{\footnotesize{ACC.}}} & \multicolumn{1}{c}{\emph{\footnotesize{ACC.}}} & \multicolumn{1}{c}{\emph{\footnotesize{ACC.}}} & \multicolumn{1}{c} {\emph{\footnotesize{Kendall's $\tau$}}}& \multicolumn{1}{c}{\emph{\footnotesize{ACC.}}}& \multicolumn{1}{c}{\emph{\footnotesize{ACC.}}}& \multicolumn{1}{c} {\emph{\footnotesize{Kendall's $\tau$}}}\\
 \hline
 \parbox[t]{2mm}{\multirow{5}{*}{\rotatebox[origin=c]{90}{General}}} & 
 \verb|GPT4| & \textbf{87.8} & 68.4 & 97.4 & 73.2 & 84.6 & \textbf{98.2} & \textbf{77.0}\\ 
&\verb|ChatGPT| & 76.8 & 86.3 & 97.4 & 59.5 & 70.6 & 85.4 & 42.6\\ 
&\verb|ERNIE-Bot| & 78.3 & 87.4 & \textbf{98.7} & \textbf{79.5} & 83.3 & 94.2 & 67.13\\ 
&\verb|SparkDesk| & 65.3 & 85.0 & \textbf{98.7} & 61.5 & 53.0 & 26.6 & 30.4\\ 
&\verb|Qwen| & 84.6 & \textbf{89.4} & 95.0 & 69.6 & \textbf{85.6} & 97.1 & 67.1\\ 
&\verb|Baichuan| & 47.6 & 56.7 & 88.5 & 22.1 & 31.2 & 32.1 & 23.9\\
& \verb|ChatGLM3| & 47.2 & 88.0 & 97.6 & 33.5 & 40.6 & 60.4 & 21.1\\
\hline 
\parbox[t]{2mm}{\multirow{3}{*}{\rotatebox[origin=c]{90}{\footnotesize{Specialized}}}} &  
\verb|HuatuoGPT| & 26.6 & \textbf{48.0} & 66.6 & 24.7 & 25.6 & 20.3 & 3.4\\ 
&\verb|BenTsao| & 27.2 & 24.6 & 24.6 & 25.2 & 4.6 & 1.0 & 18.8\\ 
&\verb|MedicalGPT| & \textbf{41.3} & 43.7 & \textbf{81.4} & \textbf{39.5} & \textbf{31.0} & \textbf{20.4} & \textbf{21.7}\\
&\verb|ChatMed| & 13.6 & 37.4 & 20.6 & 4.5 & 8.6& 2.8 & 1.5\\
 \hline
 \hline
 \multicolumn{2}{l|}{\textbf{\quad Model}} & \multicolumn{1}{c}{CS}& \multicolumn{1}{c}{FIT} & \multicolumn{1}{c}{FTT} & \multicolumn{1}{c}{BKT}& \multicolumn{1}{c}{KE}& \multicolumn{1}{c}{MC}& \multicolumn{1}{c}{WTD}\\
 \multicolumn{2}{c|}{} & \multicolumn{1}{c}{\emph{\footnotesize{ACC.}}} & \multicolumn{1}{c}{\emph{\footnotesize{ACC.}}} & \multicolumn{1}{c}{\emph{\footnotesize{ACC.}}} & \multicolumn{1}{c}{\emph{\footnotesize{ACC.}}}& \multicolumn{1}{c}
{\emph{\footnotesize{ROUGE-1}}}
 & \multicolumn{1}{c}{\emph{\footnotesize{ACC.}}}& \multicolumn{1}{c}{\emph{\footnotesize{Similarity}}}\\
 \hline
 \parbox[t]{2mm}{\multirow{5}{*}{\rotatebox[origin=c]{90}{General}}} & 
 \verb|GPT4| & 98.4& \textbf{25.0} & 12.6& 70.8 & 40.2 & 44.0 & 81.3\\ 
&\verb|ChatGPT| & 97.1 & 2.8 & 1.3 & 51.9 & 39.8& 38.9 & 80.9\\ 
&\verb|ERNIE-Bot| & \textbf{99.7} & 13.5 & 10.7 & 79.8 & \textbf{42.0} & 53.3 & \textbf{81.9}\\ 
&\verb|SparkDesk| & 95.6 & 11.7 & 4.0 & 68.7 & 28.8 & \textbf{63.5} & 81.0\\ 
&\verb|Qwen| & 99.1 & 13.9 & 13.3 & \textbf{82.4} & 39.7 & 49.2 & 80.4\\ 
&\verb|Baichuan| & 73.4 & 1.7 & \textbf{21.2} & 38.8 & 33.6 & 37.1 & 78.6\\
& \verb|ChatGLM3| & 92.0 & 9.6 & 6.8 & 46.9 & 34.4 & 45.5 & 78.8\\
\hline 
\parbox[t]{2mm}{\multirow{3}{*}{\rotatebox[origin=c]{90}{\footnotesize{Specialized}}}} &  
\verb|HuatuoGPT| & 61.2 & \textbf{13.8} & \textbf{8.6} & 22.6 & 29.3 & 23.0 & \textbf{79.2}\\ 
&\verb|BenTsao| & 25.6 & 0.6 & 0& 20.6 & 6.5& 27.6 & 75.2\\ 
&\verb|MedidalGPT| & \textbf{67.2} & 1.9 & 7.1 & \textbf{35.0} & \textbf{33.3} & \textbf{41.7} & 77.1\\
&\verb|ChatMed| & 10.9 & 2.0 & 1.2 & 9.3& 11.4 & 12.4 & 75.8\\
 \hline
 \end{tabular}
 \caption{Results of 11 LLMs with automatic metrics on the 14 core clinic scenarios of CliMedBench.}
 \label{tab: table1}
 \end{table*}

\textbf{Chinese medical LLMs underperform on this benchmark, especially where medical reasoning and factual consistency are vital}.  Comparative analysis reveals that models via APIs generally outperform others, with average scores exceeding 50. ERNIE-Bot, GPT-4, and Qwen achieve fairly similar average scores of 69.2, 69.0, and 68.5, respectively. In contrast, current medical LLMs exhibit notably inadequate performance: Even the best-performing MedicalGPT only achieves an average score of 38.7. This deficiency primarily stems from the substandard language understanding capabilities of those LLMs.

\textbf{Several Chinese LLMs} (ERNIE-Bot and Qwen) \textbf{demonstrate performance on par with GPT-4 in clinical medicine of China}, achieving scores primarily within the range of 68.5 to 69.2. This could stem from the unique treatments, expression styles, and China-manufactured pharmaceuticals, which diverge from what is encountered in the training data of GPT series models. A disparity of capabilities between these Chinese LLMs and the GPT series predominantly manifests in medical knowledge and reasoning.

Next, we will summarize the performance of LLMs with regard to particular evaluation dimensions.
Regarding clinical QA abilities, Qwen outperforms others with an average score of 88.7. However, variability in model performance across scenarios is evident, with ChatGPT achieving the highest score (97.4) on ID \#2 but not ranking among the top performers in other scenarios. GPT4 and ERNIE-Bot show exceptional reasoning capabilities, achieving average scores of 75.1 and 73.3, respectively. The notable performance disparity between general and medical-specific LLMs highlights the need for further enhancement in the reasoning ability of medical-specific LLMs. In all evaluated models, hallucinations are significantly pronounced. The FIT data is designed to trigger hallucinations by incorporating an erroneous reference. Their data sources are the same as BKT, however, model accuracy exhibits a marked reduction, plummeting from an average score of 47.3 to 8.3. This substantial decline shows the vulnerability of LLMs to uncritically adopt perspectives presented in their input, highlighting an immediate need for enhancement. Hallucinations exhibited on the WTD dataset indicate that for questions with special structures, LLMs not only need to master the knowledge points examined by the questions but also need to understand the logical relationships in the questions, which may exceed the ability of the models. The knowledge application ability of the leading general LLMs ranges from 79.8 to 82.4, suggesting a substantial reservoir of medical knowledge of these models. For the information retrieval task, MedicalGPT has significantly narrowed the disparity with leading LLMs, achieving a score of 33.3, merely 8.7 points below the top-performing ERNIE-Bot. This improvement predominantly stems from the specialized nature of its generated terminologies.

\section{Quantitive Analysis}
\begin{figure}[hbtp]
    \vspace{-6pt}
    \centering
    \includegraphics[width=7.5cm,height=5cm]{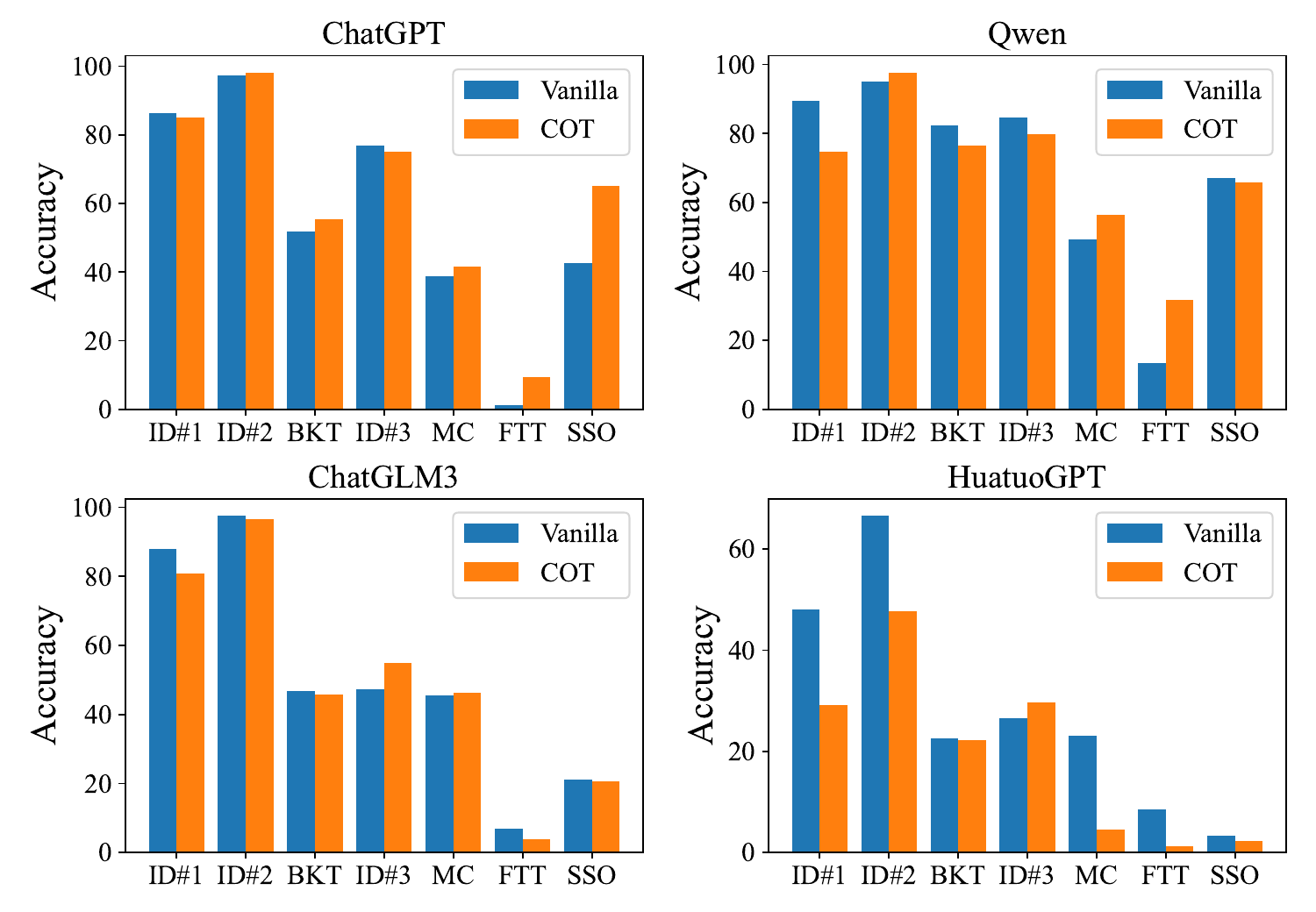}
    \caption{Accuracy comparison of four models on seven datasets using both vanilla and CoT prompts.}
    \label{fig: COT result fig}
    \vspace{-6pt}
\end{figure}

\paragraph{Chain-of-Thought}
To demonstrate the potential improvement of LLM's reasoning abilities with customized prompts, we compare the performance of four representative models using vanilla and Chain-of-Thought (CoT) prompts in Figure~\ref{fig: COT result fig}. 

We observe that the utilization of tailored CoT prompts significantly enhances model performance across seven datasets that demand higher reasoning skills. Specifically, for Qwen, there is an average accuracy increase from 62.7\% to 69.2\%. It suggests that CoT can enhance reasoning and hallucination resistance in medical contexts, as observed in the Surgical Organization, False Info Test, and Wrong Treatment datasets. Conversely, the impact of COT prompts on ChatGLM3 is minimal and it adversely affects huatuoGPT, underscoring the dependency of CoT prompt efficacy on the model's comprehension and contextual correlation proficiency. In addition, the long text of the few-shot COT prompt (on average 4.987 times longer than the vanilla prompt) is also a reason for the decrease in accuracy, as described in the following paragraph.

\begin{figure}[htbp]
  \centering
\includegraphics[width=0.8\linewidth]{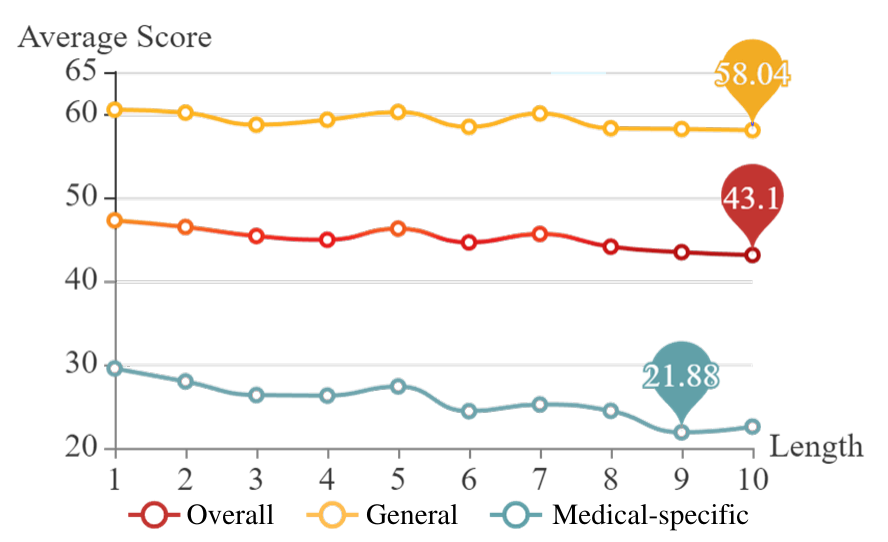}
  \caption{Performance across varying input windows, where the x-axis represents dataset segments, 1 being the shortest and 10 the longest, sorted by length.} 
  \label{fig: length influence}
  \vspace{-6pt}
\end{figure}

\paragraph{Limited Input Capacity}
We notice that EHRs frequently contain a variety of diagnostic test outcomes, records of prior treatments, and familial and social histories, often spanning multiple pages.
Consequently, the limited input capacity of many LLMs poses a challenge to their practical use in clinical scenarios.  Figure~\ref{fig: length influence} depicts performance 
comparisons across varying input windows.  

We observe a notable decrease, declining from 47.3 to 43.1, in the performance of nearly all LLMs as the length of the inputs increases, revealing that the limited input capacity is the main factor hindering their performance in clinical medicine. In addition, medical LLMs exhibit a more pronounced decline (from 29.5 to 22.6) compared to the general LLMs (from 60.5 to 58.0), suggesting that these specialized LLMs may be less capable of maintaining performance with longer inputs.

\begin{figure}[htpb]
    \vspace{-6pt}
    \centering
\includegraphics[width=0.9\linewidth]{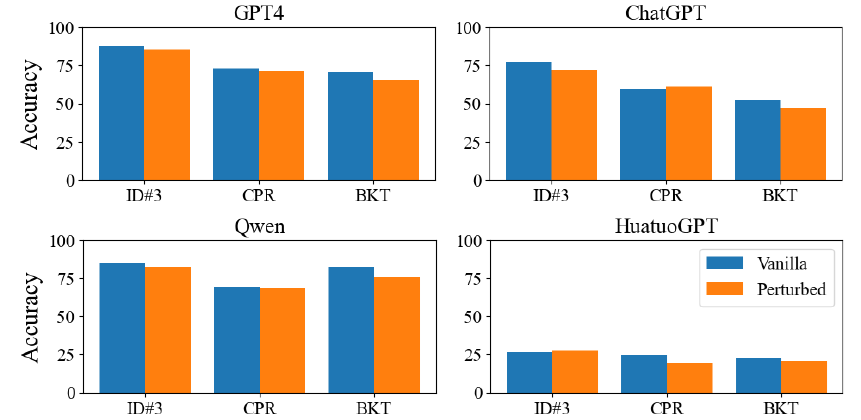}
    \caption{Robustness test of GPT4, ChatGPT, Qwen, and HuatuoGPT on different datasets.}
    \label{fig: robustness result fig}
    \vspace{-4pt}
\end{figure}

\paragraph{Robustness Test}
To conduct a robustness test, we introduce manually-crafted perturbations that comprise shape-based character conversion, homophonic substitution, simplified-to-traditional Chinese transformation, and random symbol
insertion. These perturbations cover 12\% of the characters. Figure~\ref{fig: robustness result fig} provides the robustness test results. 

We observe that with perturbations, all models exhibited a reduction in scores, ranging from 2.0 to 3.2, with particularly notable decreases observed on the basic knowledge test scenario, averaging at 4.7. This shows the significant impact of even minor disturbances on model performance despite their seemingly negligible impact on readability.

\begin{figure}[htbp]
    \vspace{-6pt}
  \centering
\includegraphics[width=1\linewidth]{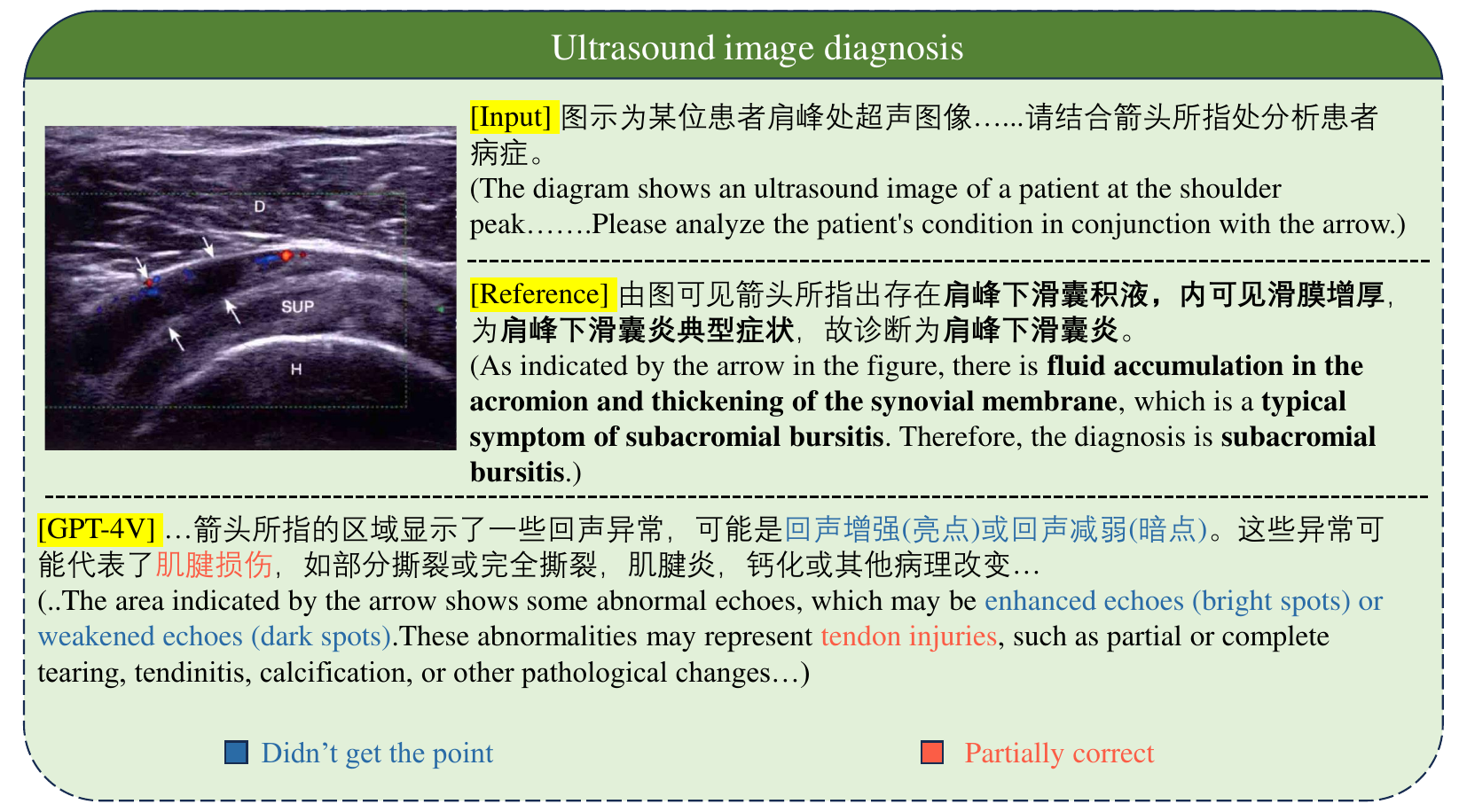}
  \caption{A multi-modal example case.}
  \label{fig: Example of VQA-pair}
    \vspace{-6pt}
\end{figure}

\paragraph{Multi-modal Capability}
To further investigate the performance of models in multi-modal settings, we have compiled a set of 92 diagnostic image pairs from medical textbooks (primarily consisting of MRI and ultrasound scans) and PathVQA~\citep{he2020pathvqa} to assess the potential of LLMs in multimodal medical diagnosis. Our evaluation primarily focuses on the representative model GPT-4V. GPT-4V does not achieve satisfying results
in these cases, and only 16.7\% of its responses are relevant to the reference. Figure~\ref{fig: Example of VQA-pair} depicts an example case, where  GPT-4V demonstrates its utility by successfully identifying inflammation in the patient's shoulder tendon through ultrasound imaging diagnostics. However, detailed but crucial diagnostic indicators within the images remained virtually indiscernible. Moreover, GPT-4V often disclaims diagnostic capability and withholds responses. Therefore, its utility in multi-modal clinical scenarios is currently limited. 

\paragraph{Inadequate Instruction Following Ability}
During the assessment, we observe that LLMs exhibit inadequate instruction following ability, a deficiency particularly conspicuous within medical-specific LLMs. Even in straightforward multiple-choice scenarios, several medical-specific LLMs, e.g., ChatMed and BenTsao, struggle to follow the given instructions to accomplish the task. Therefore, the average scores for ChatMed and Bentsao are only a quarter to a sixth of that achieved by the top-performing model. This highlights the necessity of enhancing the model's aptitude for comprehending and following diverse instructions, thereby enabling adaptation to different tasks. 

\paragraph{Potential Causes of Toxicity}
When evaluating the toxicity, we find that general LLMs, guided by safety protocols, often err on the side of caution, indiscriminately flagging and inhibiting potentially hazardous medical actions, including some that are clinically justified. Conversely, medical-specific LLMs disproportionately focus on the potential benefits of medical interventions, often neglecting potential patient-specific repercussions, e.g., advising a feverish patient who recently consumed alcohol to consider taking acetaminophen. This one-sided approach by each model type leads to suboptimal performance in toxicity assessment.

\paragraph{Lack of Innovative Thinking}
As experts pointed out in the human assessment, the responses of LLMs on CliMedBench
significantly lack innovation. To quantify this, we substitute the option of the correct answer in multiple-choice questions with ``None of the above is correct”~\citep{pal2022medmcqa}. Surprisingly, we find that this triggers a fall in accuracy to less than 10\% 
 in the false treatment test scenario for the majority of models. This suggests that, with instruction tuning, LLMs often opt for a seemingly reasonable choice from the given alternatives, potentially overlooking more precise solutions in clinical scenarios, thereby limiting the innovative capacity of medical LLMs. 
 
\section{Benchmark Validity Verification}
\label{sec:validate}
\begin{figure}[ht]
\centering
\includegraphics[width=7cm]{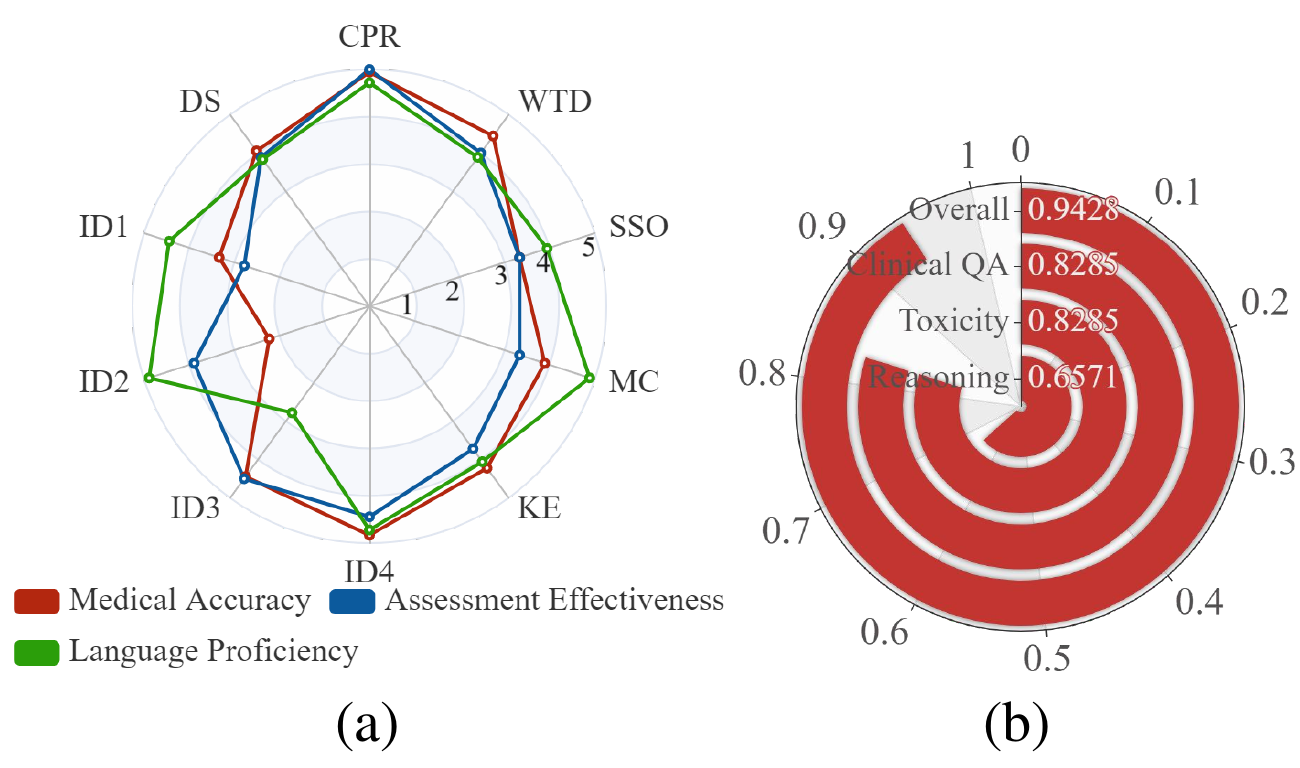}
\caption{Figure (a) depicts the assessment results of medical experts, while Figure (b) shows a noteworthy correlation between CliMedBench and MedBench.}
\label{fig: validate}
\end{figure}

To ensure the real-world applicability and validity of CliMedBench in clinical settings, the design rationale of the 14 core scenarios is meticulously aligned with actual clinical practices. For instance, the \textbf{In-hospital Diagnosis} datasets are carefully structured to adhere to established clinical pathways in healthcare settings. To further confirm the reliability of CliMedBench, we engaged medical professionals to assess our benchmark from three perspectives, including medical accuracy, assessment effectiveness, and language proficiency. Figure~\ref{fig: validate} plots the assessment results, which substantiate the quality of CliMedBench with an ``acceptable” (3 points) or higher rating.

Furthermore, we calculate the Spearman correlation between our CliMedBench and another representative benchmark, namely MedBench\footnote{https://medbench.opencompass.org.cn/leaderboard} based on other kinds of data.
This approach\footnote{https://github.com/ctlllll/understanding\_llm\_benchmark\\s?tab=readme-ov-file} allows us to conduct multidimensional evaluations that reflect both collective and discrete correlations between benchmarks. Figure~\ref{fig: validate} illustrates a robust correlation between the CliMedBench and MedBench leaderboards, with an overall Spearman correlation of 0.943 and subdivisions no less than 0.657, substantiating CliMedBench's utility and reliability as an evaluative benchmark.

\section{The Agent-based Computerized Adaptive Testing (CAT) Approach}

\begin{figure}[h]
    \vspace{-6pt}
    \centering
    \includegraphics[width=0.9\linewidth]{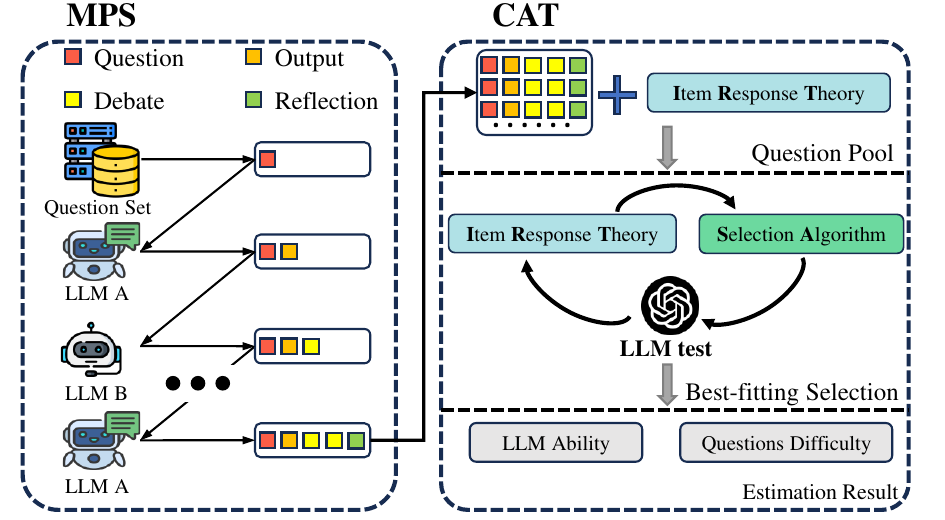}
    \caption{The workflow of Agent-based CAT.}
    \label{fig: CAT_1}
    \vspace{-4pt}
\end{figure}
During the evaluation phase, we identified two key issues: (1) Smaller LLMs struggle with exceedingly difficult questions, resulting in uniformly low accuracy and a lack of differentiation in the evaluations. (2) Certain LLMs exhibit slow GPU inference speeds or high API-related computational costs, significantly increasing benchmark testing expenses. To address this issue, we propose an agent-based CAT approach, enabling rapid assessment of model performance using CliMedBench.

\paragraph{Theoretical Basis} 
    Our approach is fundamentally rooted in the psychometric Item Response Theory (IRT). We incorporate the three-parameter logistic model (IRT-3PL), formulated as:

\begin{equation}
\vspace{-6pt}
P(X_{ij}=1|\theta_j)=c_i+(1-c_i)\frac{1}{1+e^{-a_i(\theta_j-b_i)}}
\label{eq}
\end{equation}
\noindent where $a_{i}$, $b_{i}$, and $c_{i}$ represent the discrimination, difficulty,  and the guessing factor, respectively. $\theta_j$ represents the proficiency of LLM $j$, and $P(X_{ij} = 1|\theta_j)$ is the probability that an LLM $j$ with proficiency $\theta_j$ gives a correct response to question $i$. 

\paragraph{Procedure of Agent-based CAT}
As depicted in Figure~\ref{fig: CAT_1}, our agent-based CAT consists of two main steps: Multi-Agent Based Participant Synthesis (MPS) and Computerized Adaptive Testing (CAT). MPS leverages multi-agent LLMs to synthesize data that mimics participant behavior, which is used to construct a question pool. We then use CAT to sequentially select the best-fitting questions from the question pool to evaluate 
the ability of LLMs.

Algorithm~\ref{alg: algorithm} provides the  ``generation-debate-reflection” process in MPS for data synthesis, aiming to overcome the difficulty of insufficient participant data in previous CAT~\cite{zhuang2023efficiently}. Once a sufficient number of participant behavior data is synthesized, the modeling process aligns more closely with the assumptions of IRT. Specifically, we use permutations of 5 LLMs (e.g., Bloomz and ChatGLM2) to form multi-agent based participants, serving as the examinees in the IRT process to synthesize performance-related data. 

\vspace{-4pt}
\begin{algorithm}[h]
\caption{The MPS process}
\label{alg: algorithm}
\begin{algorithmic}[1]
\State \textbf{Input:} \text{$Q$=Question data;}
\State \textbf{Output:} \text{Synthesize data}
\Function{\text{Multi-agent}}{$Q,\textrm{LLM}_1,\textrm{LLM}_2$}
    \For{$Q_i$ in $Q$}
        \State {$G \leftarrow $}{\text{$LLM_1$-Generation}}{$(Q_{i})$}
        \State {$D \leftarrow$}{\text{$LLM_2$-Debate}}{$(Q_{i},G)$}
        \State {$R \leftarrow$} {\text{$LLM_1$-Reflection}}{$(Q_{i},G,D)$}
        \State \text{DebateResults} {$\leftarrow$} \text{$(Q_{i},G,D,R)$}
    \EndFor
    \State \Return Synthesized data
\EndFunction
\end{algorithmic}
\end{algorithm}
\vspace{-4pt}

To accomplish the best-fitting selection, the CAT step includes two components that work alternately, including 
 (1) ability estimation using IRT in Eq.~\ref{eq} and (2) question selection via  Fisher information.  
 
\begin{table}[h]
 \vspace{-4pt}
 \centering
    \begin{tabular}{l|c}
    \hline
    \textbf{Method} & \textbf{Accuracy}  \\ \hline
    CAT~\cite{zhuang2023efficiently} & 32.10 \\ 
    Our Agent-based CAT & 40.74   \\ \hline
    \end{tabular}
    \caption{Comparison to other CAT method.}
    \label{tab: test}
 \vspace{-4pt}
\end{table}


\begin{table*}[t]

 \centering
    \scalebox{0.90}{\begin{tabular}{cccccc}
    \hline
    \textbf{Benchmark} & \textbf{Language} & \textbf{Data Source} & \textbf{Coverage} & \textbf{Size}\\ \hline
    CliMedBench & Chinese & \makecell{EHR(major), NMLEC,\\and Books} & 14 Core Scenarios & 33.7k\\ 
    CMExam & Chinese & CNMLE & Exam QA & 60k\\
    CMB & Chinese & \makecell{Exams and Chinese Medical\\Question Database} & Exam QA & 281k\\ 
    MLEC-QA & Chinese & NMLEC and Books & Exam QA & 136k\\
    RJUA-QA & Chinese & EHR & Urology & 2.1k\\ 
    MedBench & Chinese & CNMLE and NMLEC & Exam QA & 40k\\ 
    NLPEC & Chinese & NLPEC & Exam QA & 21.7k\\ 
    MultiMedQA & English & Existing datasets & - & 18.8k\\
    PubMedQA & English & PubMed & Biomedical paper QA & 1k labeled\\ 
    emrQA & English & EHR & EHR QA & 400k\\ \hline
    \end{tabular}}
    \caption{Contrasts between the CliMedBench and existing benchmarks.}
    \label{tab: related work}
\end{table*}

\begin{figure}[ht]
\centering
\includegraphics[width=7.8cm]{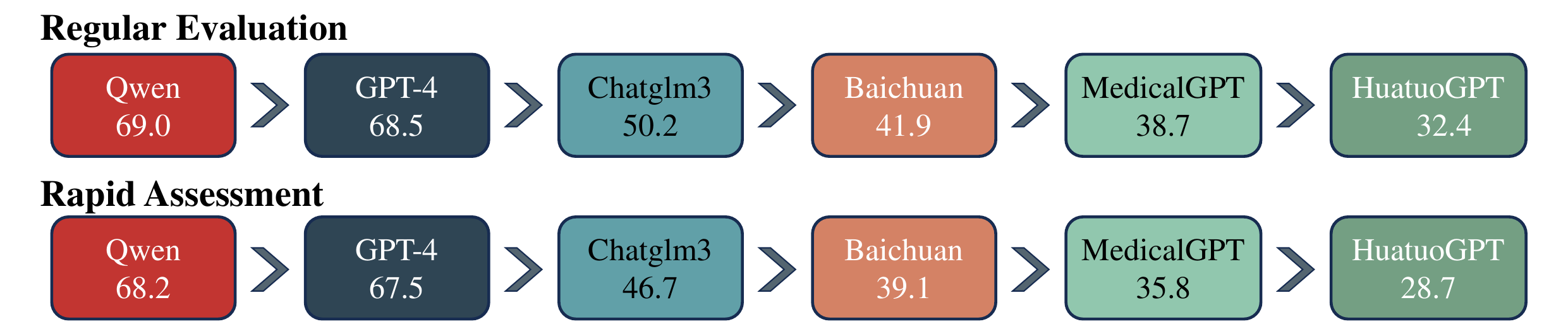}
\caption{Comparisons between regular CliMedBench evaluation and our agent-based
CAT rapid assessment.}
\label{fig: 200vs20k}
\vspace{-8pt}
\end{figure}

\paragraph{Results}
We select 243 questions from CliMedBench to conduct a rapid assessment using our agent-based Computerized Adaptive Testing (CAT). To validate the effectiveness of our agent-based CAT rapid assessment, we compare its results with the regular CliMedBench evaluation, which involved 33,735 instances as proposed in Section~\ref{benchmark}. This comparison is illustrated in Figure~\ref{fig: 200vs20k}. Our observations indicate a consistency in the relative rankings of LLMs derived from the two evaluation methods, validating the effectiveness of using a limited set of questions to gauge model ability.

Table~\ref{tab: test} compares the accuracy of our agent-based CAT with that of the representative previous work by \citet{zhuang2023efficiently}. We observe a relative performance increase of 26.9\%, demonstrating the efficacy of MPS in synthesizing sufficient data that better aligns with the IRT assumption.


\section{Related Work}
MultiMedQA~\citep{singhal2023large} and PubMedQA~\citep{jin-etal-2019-pubmedqa} are effective benchmarks based on QA tasks for evaluating the medical abilities of LLMs. Large-scale EHR-based benchmarks such as emr-QA~\citep{pampari-etal-2018-emrqa} have addressed the deficiency in clinical QA, but language discrepancies preclude their direct applicability for evaluating Chinese medical LLMs. The assessment of LLMs for Chinese medical proficiency has traditionally relied on benchmarks derived from multiple-choice and generative question-answering formats, utilizing resources like exam questions, textbooks, and doctor-patient interactions. Chinese benchmarks, including CMExam~\citep{liu2024benchmarking}, CMB~\cite{wang-etal-2024-cmb}, MLEC-QA~\citep{li2021mlec}, and MedQA~\cite{jin2021disease}, primarily source their data from exams such as CNMLE and NMLEC. Despite their comprehensive analysis, these benchmarks are disconnected from actual medical practice due to their lack of real-world medical case data, and the challenge of ensuring quality control and avoiding data pollution grows proportionally with the involved volume of data~\citep{li2023huatuo}. ~\citet{li-etal-2020-towards} collected more than 21k multi-choice questions from the National Licensed Pharmacist Examination in China, but only the test set of the dataset has been released.

Finally, we contrast our work with other lines of work sharing seemingly similar goals. MedBench~\citep{cai2024medbench} is an exhaustive benchmark designed for the domain of Chinese medical QA, it utilizes exam questions and synthetical EHRs to evaluate the LLMs' exam-solving capabilities in different areas, rather than actual clinical skills. In contrast, our benchmark extends this framework across 14 diverse medical scenarios. Furthermore, despite MedBench providing preliminary empirical analysis, it lacks in-depth qualitative analyses of the model's performance. RJUA-QA~\citep{lyu2023rjua} creates high-quality medical datasets to evaluate clinical reasoning based on EHRs and clinical cases. However, it is restricted to urology, offering limited insight into the broader medical capabilities of LLMs. Table~\ref{tab: related work} delineates the contrasts between the CliMedBench and existing benchmarks from the perspectives of language, data sources, coverage, and size.

\section{Conclusion and Discussions}
This paper introduces CliMedBench, a robust benchmark derived from real medical cases that comprises 33,735 questions across 14 core medical scenarios assessing LLMs' ability across six dimensions. Evaluating diverse LLMs reveals their suboptimal performance, especially where medical reasoning and factual consistency are vital, underscoring the need for advances in clinical knowledge and diagnostic accuracy. We also conducted a comprehensive qualitative analysis of the experimental outcomes and made several novel insights. Simultaneously, we proposed the agent-based CAT approach, which enables rapid assessment with minimal problem sets. 


\section{Limitations and Ethical Issues}
Protected Health Information (PHI) encompasses data related to an individual's health status, healthcare provision, or payment for healthcare services, which is generated or amassed by a Covered Entity or its Business Associate. PHI typically undergoes de-identification to safeguard individual privacy prior to the dataset's publication. 
CliMEdBench is a dataset derived primarily from real-world medical cases and the Chinese National Physician Qualification Examination. All EHRs and codes have been doubly de-identified by ethics committees and experts according to the guidance and have passed the ethical review of our partner hospitals before submission. However, such real-world data may suffer from noise. 
This stems from two main sources: (i) erroneous data input by medical personnel during recording or formatting error during data retrieval, and (ii) inaccuracies introduced in automatic information extraction. 
Users should exercise caution regarding data reliability in light of these limitations. 
In future work, extensive validation by medical experts will be conducted to ensure the correctness of all data.
Our project has been conducted in collaboration with relevant medical centers with proper approval of all data sharing.
Due to legal restrictions, our data is available for research purposes only. Researchers can contact us with the research objectives and intended use of the data.
We ensure full compliance with applicable laws and ethical guidelines during data collection and use, all information in medical cases has been desensitized to ensure that no personal information related to patients or medical personnel is leaked. 

\bibliography{custom}

\end{document}